\documentclass[runningheads]{llncs}
\usepackage{graphicx}
\usepackage{amsmath}
\usepackage{booktabs} % For pretty tables
\usepackage{caption} % For caption spacing
\usepackage{subcaption} % For sub-figures
\usepackage{graphicx}
\usepackage{pgfplots}
\usepackage[all]{nowidow}
\usepackage[utf8]{inputenc}
\usepackage{tikz}
\usetikzlibrary{er,positioning,bayesnet}
\usepackage{multirow,bigdelim}
\usepackage{multicol}
\usepackage{algpseudocode,algorithm,algorithmicx}
\usepackage{hyperref}
\usepackage[inline]{enumitem} % Horizontal lists
\pgfplotsset{compat=1.14}

\newcommand\KL{\text{\bf{KL}}}

\begin{document}
\title{Deep Active Inference for Pixel-Based Discrete Control: Evaluation on the Car Racing Problem\thanks{2nd International Workshop on Active Inference IWAI2021, European Conference on Machine Learning (ECML/PCKDD 2021)}}
\titlerunning{Deep Active Inference: Evaluation on the Car Racing Problem}
\author{Niels van Hoeffelen \and Pablo Lanillos}
\authorrunning{N.T.A. van Hoeffelen, P. Lanillos}
\institute{Department of Artificial intelligence\\
Donders Institute for Brain, Cognition, and Behaviour\\
Radboud University\\
Montessorilaan 3, 6525HR Nijmegen, the Netherlands\\
\email{niels.vanhoeffelen@ru.nl}\\ 
\email{p.lanillos@donders.ru.nl}}
\maketitle              % typeset the header of the contribution
\begin{abstract}
Despite the potential of active inference for visual-based control, learning the model and the preferences (priors) while interacting with the environment is challenging. Here, we study the performance of a  deep active inference  (dAIF)  agent on  OpenAI’s car racing benchmark, where there is no access to the car’s state. The agent learns to encode the world’s state from high-dimensional input through unsupervised representation learning. State inference and control are learned end-to-end by optimizing the expected free energy. Results show that our model achieves comparable performance to deep Q-learning. However, vanilla dAIF does not reach state-of-the-art performance compared to other world model approaches. Hence, we discuss the current model implementation's limitations and potential architectures to overcome them.

\end{abstract}
\section{Introduction}
Learning from scratch which actions are relevant to succeed in a task using only high-dimensional visual input is challenging and essential for artificial agents and robotics. Reinforcement learning (RL)~\cite{sutton2018reinforcement} is currently leading the advances in pixel-based control, e.g., the agent learns an action policy that maximizes the accumulated discounted rewards. Despite its dopamine biological inspiration, RL is far from capturing the physical processes happening in the brain. We argue, that prediction in any form (e.g., visual input, muscle feedback or dopamine) may be the driven motif of the general learning process of the brain~\cite{lanillos2021neuroscience}. Active inference~\cite{friston2010action,da2020active}, a general framework for perception, action and learning, proposes that the brain uses hierarchical generative models to predict incoming sensory data \cite{friston2005theory} and tries to minimize the difference between the predicted and observed sensory signals. This difference, mathematically described as the variational free energy (VFE), needs to be minimized to generate better predictions about the world that causes these sensory signals~\cite{friston2010action,millidge2020deep}. Through acting on its environment, an agent can affect sensory signals to be more in line with predicted signals, which in turn leads to a decrease of the error between observed and predicted sensory signals. 

AIF models have shown great potential in low-dimensional and discrete state spaces. To work in higher-dimensional state spaces, deep active inference (dAIF) has been proposed, which uses deep neural networks for approximating probability density functions (e.g., amortised inference)~\cite{millidge2020deep,tschantz2020scaling,ueltzhoffer2018deep,ccatal2019bayesian,sancaktar2020end}. Interestingly, dAIF can be classified as a generalization of the world models approach~\cite{ha2018world} and can incorporate reward-based learning, allowing for a direct comparison to RL methods. Since the first attempt of dAIF~\cite{ueltzhoffer2018deep}, developments have happened concurrently in adaptive control~\cite{sancaktar2020end,meo2021multimodal} and in planning, exploiting discrete-time optimization, e.g., using the expected free energy~\cite{parr2019generalised,tschantz2020reinforcement}. In \cite{millidge2020deep}, a dIAF agent was tested on several environments in which the state is observable (Cartpole, Acrobot, Lunar-lander). In \cite{fountas2020deepMC}, a dAIF agent using Monte-Carlo sampling was tested on the Animal-AI environment. In \cite{ccatal2020learning}, dAIF tackled the mountain car problem and was also tested on OpenAI's car racing environment. For the car racing environment, they trained the dAIF agent on a handful of demonstration rollouts and compared it to a DQN that interacted with the environment itself. Their results showed that DQN needs a lot more interaction with the environment to start obtaining rewards compared to dAIF trained on human demonstrations of the task. Relevant for this work, in \cite{van2020deep}, a dAIF agent solved the Cartpole environment as both MDP and as POMPD instances, training the agent on just visual input. 

%To extend the time horizon which is taken into account, minimization of the Expected-free-energy (EFE) has been proposed \cite{parr2019generalised,millidge2020deep}. %Where for each possible action we take into account the possible future trajectories and calculate the EFE up to some time k.

%% Closing intro: What happens in this paper
In this paper, we study a dAIF agent\footnote{The code can be found at \url{https://github.com/NTAvanHoeffelen/DAIF_CarRacing} } based on the proposed architectures in \cite{millidge2020deep,van2020deep} for a more complex pixel-based control POMDP task, namely the OpenAI's Car Racing environment~\cite{openCar}, and discuss its advantages and limitations compared to other state-of-the-art models. The performance of the dAIF agent was shown to be in line with previous works and on-par with deep Q-learning. However, it did not achieve the performance of other world model approaches~\cite{carLeaderboard}. Hence, we discuss the reasons for this, as well as architectures that may help to overcome these limitations.

\section{Deep Active Inference Model}

\begin{figure} [hbtp!]
        \centering
        \includegraphics[width=1\linewidth]{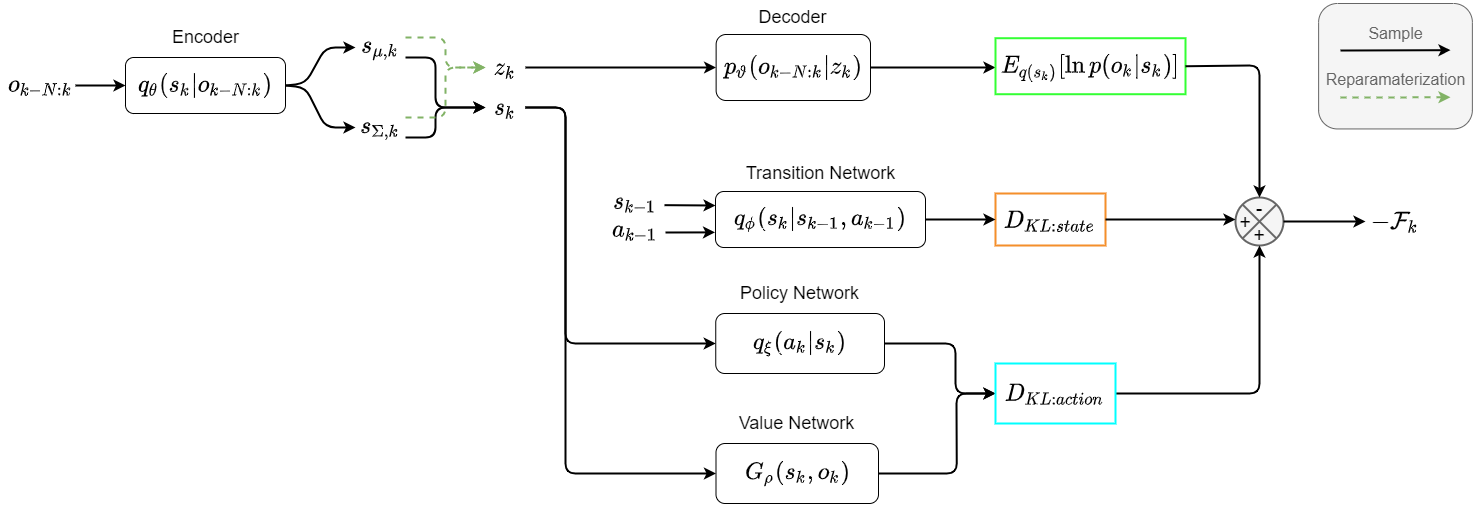}
        \caption{Deep Active Inference architecture. Five deep artificial neural networks are used to model the observation encoding and decoding, the state transition, the policy and the EFE values. The architecture is trained end-to-end by optimizing the expected variational free energy.}
        \label{fig:dAIF-diagram}
\end{figure}

%%%%%%%%%%%%%%%%%%%% MODEL %%%%%%%%%%%%%%%%%%%%%%%%
The dAIF architecture studied is based on~\cite{van2020deep} and described in Fig.~\ref{fig:dAIF-diagram}. It makes use of five networks to approximate the densities of Eq.~(\ref{eq:f2}): observation (encoding and decoding), transition, policy, and value. The full parameter description of the networks can be found in the Appendix~\ref{appendix:parameters}.

\subsubsection{Variational Free Energy (VFE).} AIF agents infer their actions by minimizing the VFE expressed at instant $k$ (with explicit action 1-step ahead) as: 
\begin{align}
\mathcal{F}_k = - \KL[q(s_k,a_k)\mid \mid p(o_k,s_{0:k},a_{0:k})]
\label{eq:f1}
\end{align}
Where $s_k,o_k,a_k$ are the state, the observation and action respectively, $q(s_k,a_k)$ is the recognition density, and $p(o_k,s_{0:k},a_{0:k})$ the generative model. Under the Markov assumption and factorizing~\cite{millidge2020deep,van2020deep}, Eq.~(\ref{eq:f1}) can be rewritten as:
%\begin{align}
%p(o_t,s_t,a_t,s_{t-1},a_{t-1}) = p(o_t|s_t)p(a_t|s_t)p(s_t|s_{t-1},a_{t-1})q(s_{t-1}, a_{t-1}) \nonumber
%\end{align}

%We can also factorize the recognition distribution, following the laws of probability we get:
%\begin{align}
%q(s,a) = q(a_t|s_t)q(s_t)
%\nonumber
%\end{align}

%This allows us to write Eq. 2.1 as:
%\begin{align}
%F_t = - D_{KL}[q(a_t|s_t)q(s_t)\mid \mid p(o_t|s_t)p(a_t|s_t)p(s_t|s_{t-1},a_{t-1})]
%\end{align}

%Were we can omit $q(s_{t-1}, a_{t-1})$ because is not dependent on any variational parameters of $q(s_t, a_t)$. Giving us:

\small
\begin{align}
\mathcal{F}_k = {} & \underbrace{E_{q(s_k)}[\ln{p(o_k|s_k)}]}_{\text{sensory prediction}} - \underbrace{\KL[q(s_k)\mid \mid p(s_k|s_{k-1},a_{k-1})]}_{\text{state prediction}}- \underbrace{\KL[q(a_k|s_k)\mid \mid p(a_k|s_k)]}_{\text{action prediction}}
\label{eq:f2}
\end{align}
\normalsize

%%%%%%%%%%%%%%%%%%%%%%%% OBSERVATION NET %%%%%%%%%%%%%%%%%%%%%%%
\subsubsection{Sensory prediction.} The observation network predicts the observations---first term in Eq.~(\ref{eq:f2})---and encodes the high-dimensional input to states $q_\theta(s_k|o_{k-N:k})$ and decodes latent spaces to observations $p_\vartheta(o_{k-N:k}|z_k)$. It can be implemented with a variational autoencoder, where latent representation is described as a multivariate Gaussian distribution with mean $s_\mu$ and variance $s_\Sigma$. The latent space $z_k$ is obtained using the reparametrisation trick. To train the network, we use the binary cross-entropy and the KL regularizer to force the latent space to be Gaussian:
\begin{equation}
\begin{split}
L_{o,k} &= -E_{q_\theta(s_k|o_{k-N:k})}[\ln{p_\vartheta(o_{k-N:k}|z_k)}] \\
&= BCE(\hat{o}_{k-N:k}, o_{k-N:k}) -\frac{1}{2} \sum(1 + \ln{s_{\Sigma,k}} - s_{\mu,k}^2 - s_{\Sigma,k}) \\
\end{split}
\end{equation}

%%%% Is this needed? %%%%%%%%%%%%%%%%%%%%%%%%%%%%%%%%%%%%%%%

% The first term of Eq. 2.3 can be split even further into an entropy component and a Kullback Leibler divergence:

% \begin{align}
% -E_{q(s_t)}[\ln{p(o_t|s_t)}] &= H(q(s_t)) + D_{KL}[q(s_t)\mid \mid p(o_t|s_t)]
% \end{align}

% Filing in the network distributions then gives us:
% \label{eq 4}
% \begin{multline} 
%     -E_{q_\theta(s_t|o_{t-7:t})}[\ln{p_\vartheta(o_{t-7:t}|z_t)}] = H(q_\theta(s_t|o_{t-7:t}))\\
%     + D_{KL}[q_\theta(s_t|o_{t-7:t})\mid \mid p_\vartheta(o_{t-7:t}|z_t)]
% \end{multline}

% To capture Eq. 2.4, the loss function of the Variable Autoencoder is used, which consists of a reconstruction term and a regularization term. The reconstruction is evaluated using binary cross-entropy (BCE). The regularization term is expressed as a Kullback-Leibler divergence.\\

% The entropy component of Eq. 2.4 is replaced with the BCE of the predicted and actual observations: 
% \begin{align}
%  H(q_\theta(s_t|o_{t-7:t})) = BCE(\hat{o}_{t-7:t}, o_{t-7:t})
% \end{align}

% The KL-divergence can be rewritten because we obtain the mean ($s_\mu$) and the variance ($s_\Sigma$) from the encoder network of the VAE: 
% \begin{align}
%     D_{KL}[q_\theta(s_t|o_{t-7:t})\mid \mid p_\vartheta(o_{t-7:t}|z_t)] = -\frac{1}{2} \sum(1 + \ln{s_{\Sigma,t}} - s_{\mu,t}^2 - s_{\Sigma,t})
% \end{align}

%%%%%%%%%%%%%%%%%%%%%%% TRANSITION NET %%%%%%%%%%%%%%%%%%%%%%%%%

\subsubsection{State prediction.} The transition network models a distribution that allows the agent to predict the state at time $k$ given the state and action at time $k - 1$, where the input state consists of both the mean $s_\mu$ and variance $s_\Sigma$. Under the AIF approach this is the difference between the state distribution (generated by the transition network) and the actual observed state (from the encoder): $\KL[q(s_k)\mid \mid p(s_k|s_{k-1},a_{k-1})]$. For the sake of simplicity, we define a feed-forward network that computes the maximum-a-posteriori estimate of the predicted state $\hat{s}_k = q_\phi(s_{k-1},a_{k-1})$ and train it using the mean squared error:
\begin{equation}
    MSE(s_{\mu,k},q_\phi(s_{k-1},a_{k-1}))
\end{equation}

%%%%%%%%%%%%%%%%%%%%%%% POLICY NET %%%%%%%%%%%%%%%%%%%%%%%%%%%%%%

\subsubsection{Action prediction.} We use two networks to evaluate action: the policy and value network. The action prediction part of Eq.~(\ref{eq:f2}) is the difference between the model's action distribution and the ``optimal" true distribution. It is a KL divergence, which can be split into an energy and an entropy term:
\begin{equation}
\begin{split}
\KL[q(a_k|s_k)\mid \mid p(a_k|s_k)] &= - \sum_{a} q(a_k|s_k) \ln{\frac{p(a_k|s_k)}{q(a_k|s_k)}}\\
&= \underbrace{- \sum_{a} q(a_k|s_k) \ln{p(a_k|s_k)}}_{\text{energy}} \underbrace{- \sum_{a} q(a_k|s_k) \ln{q(a_k|s_k)}}_{\text{entropy}}
\end{split}
\end{equation}

The policy network models the distribution over actions at time $k$ given the state at time $k$ $q_\xi(a_k|s_k)$. It is implemented as a feed-forward neural network that returns a distribution over actions given a state.

The value network computes the Expected Free Energy (EFE)~\cite{parr2019generalised,millidge2020deep,van2020deep} which is used to model the true action posterior $p(a_k|s_k)$. As the true action posterior is not exactly known, we assume that prior belief makes the agent select policies that minimize the EFE. We model the distribution over actions as a precision-weighted Boltzmann distribution over the EFE~\cite{parr2019generalised,millidge2020deep,fountas2020deepMC,schwartenbeck2019computational}: 
\begin{align}
    p(a_k|s_k) = \sigma(-\gamma G(s_{k:N}, o_{k:N}))
\end{align}
where $G(s_{k:N}, o_{k:N})$ is the EFE for a set of states and observations up to some future time $N$. As we are dealing with discrete time steps, it can be written as a sum over these time steps:
\begin{align}
    G(s_{k:N}, o_{k:N}) = \sum^N_k{G(s_k,o_k)}
\end{align}

\noindent The EFE is evaluated for every action, because of this we implicitly conditioned on every action~\cite{millidge2020deep}. We then define the EFE of a single time step as\footnote{The full derivation can be found in Appendix~\ref{appendix:EFEderivation}}:
\setlength{\jot}{8pt}
\begin{equation}
\begin{split}
G(s_k, o_k) &= \KL[q(s_k) \mid \mid p(s_k, o_k)] \\ 
&\approx - \ln{p(o_k)} + \KL[q(s_k) \mid \mid q(s_k|o_k)]\\
&\approx -r(o_k) + \KL[q(s_k) \mid \mid q(s_k|o_k)] 
\end{split}
\end{equation}

The negative log-likelihood (or surprise) of an observation $-\ln{p(o_t)}$, is replaced by the reward $-r(o_k)$~\cite{millidge2020deep,friston2012active,van2020deep}. As AIF agents act to minimize their surprise, by replacing the surprise with the negative reward, we encode the agent with the prior that its goal is to maximize reward. This formulation needs the EFE computation for all of the possible states and observations up to some time $N$, making it computationally intractable. Tractability has been achieved through bootstrapping \cite{millidge2020deep,van2020deep} and combining Monte-Carlo tree search and amortized inference~\cite{fountas2020deepMC}. Here we learn a bootstrapped estimate of the EFE. The value network is used to get an estimate of the EFE for all of the possible actions. It is modelled as a feed-forward neural network $G_{\rho}(s_k, o_k) = f_{\rho}(s_k)$.

To train the value network, we use another bootstrapped EFE estimate which uses the EFE of the current time step and a $\beta \in ( 0,1 ]$ discounted value-net estimate of the EFE under $q(a_{k+1}|s_{k+1})$ for the next time step:
\begin{equation}
\begin{split}
    \hat{G}(s_k, o_k) = -r(o_k) + \KL[q(s_k) \mid \mid q(s_k|o_k)] + \beta E_{q(a_{k+1}|s_{k+1})}[G_{\rho}(s_{k+1}, o_{k+1})]
\end{split}
\end{equation}

Using gradient descent, we can optimize the parameters of the value network by computing the MSE between $G_{\rho}(s_k, o_k)$ and $\hat{G}(s_k, o_k)$:

\begin{align}
    L_{f_{\rho},k} = MSE(G_{\rho}(s_k, o_k), \hat{G}(s_k, o_k))
\end{align}

\noindent In summary, with our implementation, the VFE loss function becomes:
\begin{equation}
\begin{split}
     - \mathcal{F}_k = & BCE(\hat{o}_{k-N:k}, o_{k-N:k}) -\frac{1}{2} \sum(1 + \ln{s_{\Sigma,k}} - s_{\mu,k}^2 - s_{\Sigma,k})\\
           & + MSE(s_{\mu,k},q_\phi(s_{k-1},a_{k-1}))\\
           & + \KL[q_\xi(a_k|s_k)\mid \mid \sigma(-\gamma G_{\rho}(s_k, o_k))]
\end{split}
\end{equation}

\section{Experimental Setup} 
\label{experiment_setup}

\begin{figure}[hbtp!]
	\centering
	\begin{minipage}{0.4\textwidth}
	\subfloat[CarRacing-v0]{
		\centering		\includegraphics[width=0.6\columnwidth]{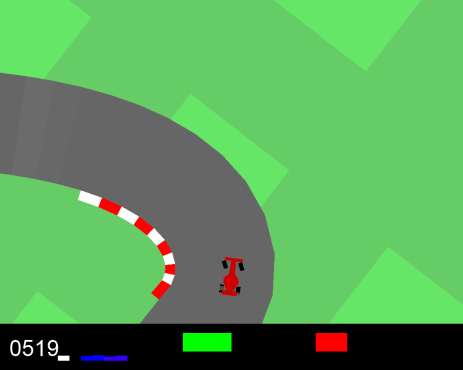}		\label{img:carracing:env}
	}\\
	\subfloat[Preprocessed input]{
	        \includegraphics[width=0.6\columnwidth]{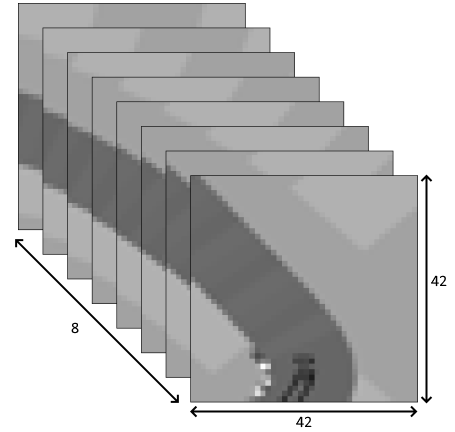}
	  \label{img:carracing:input}
	}
	\end{minipage}%
	\begin{minipage}{0.5\textwidth}
	\subfloat[Discrete Actions]{
            \centering
                 \begin{tabular}{|c|c|}
                \hline
                action & values \\
                \hline
                do nothing & [0, 0, 0] \\
                steer sharp left & [-1, 0, 0] \\
                steer left & [-0.5, 0, 0] \\
                steer sharp right & [1 ,0, 0] \\
                steer right & [0.5, 0, 0]\\
                accelerate 100\% & [0, 1, 0] \\
                accelerate 50\% & [0, 0.5, 0] \\
                accelerate 25\% & [0, 0.25, 0] \\
                brake 100\% & [0, 0, 1] \\
                brake 50\% & [0, 0, 0.5] \\
                brake 25\% & [0, 0, 0.25] \\
                \hline
            \end{tabular}
            \label{tab:discrete_actions_11}
    }
    \end{minipage}%
\end{figure}

We evaluated the algorithm on OpenAI's CarRacing-v0 environment~\cite{openCar} (Fig.~\ref{img:carracing:env}). It is considered a Partial Observable Markov Decision Process (POMDP) problem as there is no access to the state of the agent/environment. The input is the top-view image ($96 \times 96$ RGB) of part of the racing track centred on the car. The goal of this 2D game is to maximize the obtained reward by driving as fast and precise as possible. A reward of +1000/N is received for every track tile that is visited, where N is the total number of tiles (placed on the road), and a reward of -0.1 is received for every frame that passes. Solving the game entails that an agent scores an average of more than 900 points over 100 consecutive episodes. By definition, an episode is terminated after the agent visited all track tiles, the agent strayed out of bounds of the environment, or when 1000 time steps have elapsed. Every episode, a new race track is randomly generated.

The agent/car has three continuous control variables, namely steering $[-1, 1]$ (left and right), accelerating $[0, 1]$, and braking $[0, 1]$. The action space was discretized into $11$ actions, similarly to~\cite{zhang2020deep,slik2019deep,van2018advantage}, described in Table \ref{tab:discrete_actions_11}.

\section{Results}
\begin{figure} [hbtp!]
        \centering
        \includegraphics[width=0.9\linewidth]{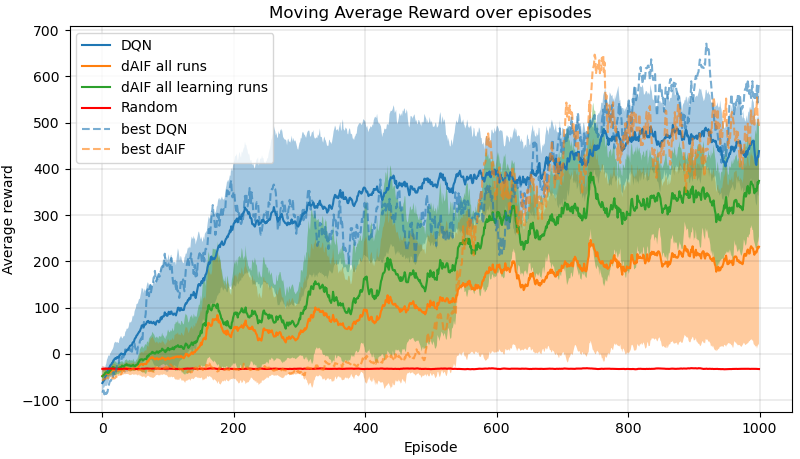}
        \caption{Moving average reward (MAR) comparison for OpenAI's CarRacing-v0. $MAR_e = 0.1CR_e + 0.9MAR_{e-1}$, where $CR_e$ is the cumulative reward of the current episode. In green (solid line), the mean of the dAIF runs that were able to learn a policy, and in orange (dashed line), the best of all training runs.}
        \label{fig:results}
\end{figure}

We compared our dAIF implementation with other state-of-the-art algorithms. First, Fig. \ref{fig:results} shows the average reward evolution while training for our dAIF architecture, Deep-Q learning (DQN)~\cite{mnih2015human} (our implementation) and a random agent. Second, Table~\ref{tab:avg_rewards} shows the average reward over 100 consecutive episodes for the top methods in the literature. The average reward performance test and reward per episode for DQN and dAIF are provided in Appendix~\ref{appendix:avgreward}.
\vspace{-0.4cm}
%\vspace{-0.6cm}
%\newpage

For the dAIF and the DQN implementations, observations are first preprocessed by removing the bottom part of the image. This part contains information about the accumulated rewards of the current episode, the car's true speed, four ABS sensors, steering wheel position, and gyroscope. Afterwards, the image is grey-scaled and reshaped to a size of $42\times42$. The input is defined as a stack of $k$ to $k-N$ observations to provide temporal information by allowing the encoding of velocity and steering. We use experience replay~\cite{Lin1992ReinforcementLF} with a batch size of 250 and memory capacity of 300000 and 100000 transitions for DQN and dAIF respectively, and make use of target networks~\cite{mnih2015human} (copy of the policy network for DQN and value network for dAIF) with a freeze period of 50 time steps.

The dAIF agent makes use of a pre-trained VAE which was frozen during training. Following a similar procedure as~\cite{ccatal2020learning}, the VAE was pre-trained on observations collected by having a human play the environment for 10000 time steps.
\begin{table} [hbtp!]
\centering
\caption{Average rewards for CarRacing-v0}
\label{tab:avg_rewards}
\begin{tabular}{cc}
\hline
Method       & Average Reward             \\
\hline
DQN (our implementation) & 515 $\pm$ 162 \\
dAIF (our implementation) & 494 $\pm$ 241                           \\
A3C (Continuous)~\cite{MinJ.A3C} & 591 $\pm$ 45  \\
A3C (Discrete)~\cite{M.khan.A3Cdisc} & 652 $\pm$ 10 \\
Weight Agnostic Neural Networks~\cite{gaier2019weight} & 893 $\pm$ 74     \\
GA~\cite{risi2019deep} & 903 $\pm$ 72 \\ 
World models~\cite{ha2018world} & 906 $\pm$ 21  \\
             &                           
\end{tabular}
\end{table}

\section{Discussion}
\label{sec:discussion}
The dAIF implementation described in this paper has shown to reach performance on par with Deep Q-learning. However, there are some remarks. First, it showed a slower learning curve as described in previous works~\cite{van2020deep}, due to the need to learn the world model. Second, we identified some runs where the system was not able to learn---See Fig.~\ref{fig:results} orange solid line. These runs drag down the average performance. Finally, it has failed to reach state-of-the-art performance when comparing to other world model approaches---See Table~\ref{tab:avg_rewards}. Here we discuss the limitations of the current implementation and alternative architectures to overcome the challenge of learning the preferences in dAIF approaches.
\subsubsection{Observation and transition model.} 
\begin{figure} [hbtp!]
        \centering
        \includegraphics[width=1\linewidth]{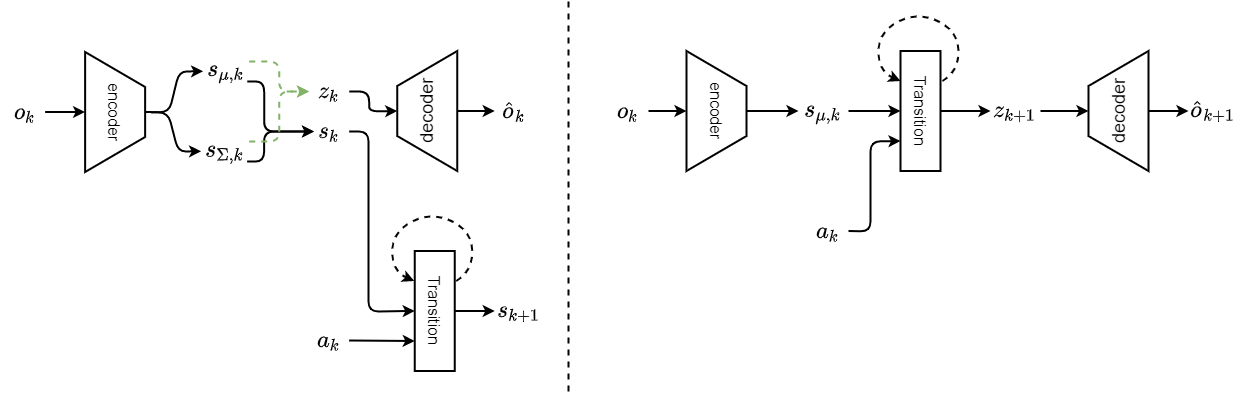}
        \caption{Transition network outside of the observation network (left) and the transition network in between the encoder and decoder (right).}
        \label{fig:alternative_architectures}
\end{figure}
Our implementation does not fully exploit temporal representation learning, such as other models that use recurrent neural networks (RNN). Figure \ref{fig:alternative_architectures} describes two architectures to learn the encoding and the transitioning of the environment. Figure \ref{fig:alternative_architectures} left shows the autoencoding and transition model implemented in this work. This architecture is similar to the successful world models~\cite{ha2018world}, but while we used a simple feed-forward network, they modelled the state-transition with a mixed density network RNN. Interestingly, dAIF also permits the alternative architecture described in Fig.~\ref{fig:alternative_architectures} right (also proposed in~\cite{noel2021online}), in which the network learns to predict future observations. Here the transition network is part of the autoencoding. By incorporating the transition network in the structure of the observation network, we avoid the need for the dual objectives: perceptual reconstruction and dynamics learning. Preliminary testing did not show any improvements.  Future work could involve more extensive testing to uncover possible performance improvements.

\subsubsection{Dependency of input space}
The performance of dAIF has shown a strong dependency on the learning of the observation model. Different image pre-processing methods would lead to improvements of more than $50\%$ in the agent performance, as shown in other DQN implementations in the literature. Testing showed that without a pre-trained observation network, the model was unable to learn consistently and rarely showed performance that would suggest an indication of task comprehension. By using a pre-trained observation network learning occurred in 6 out of the 10 runs. To produce a proper action-centric representation, likelihood, transition and control should be learnt concurrently. However, parameters uncertainty may be tackled as the models are being learnt. The current implementation uses static values for the networks learning rates, future testing could investigate different variable learning rates for each network or decaying dropout temperature.

\subsubsection{Bootstrapping of the policy and the value.} Estimating both the policy and the value from state encoding has shown end-to-end issues when we do not pre-train the observation model. In particular, 1-step ahead action formulation in conjunction with bootstrapping might not capture a proper structure of the world, which is needed to complete the task, even if we use several consecutive input images to compute the state. N-step ahead observation optimization EFE formulations, as proposed in \cite{fountas2020deepMC,tschantz2020reinforcement,noel2021online}, may aid learning. Particularly, when substituting the negative log surprise by the rewards, the agent might loose the exploratory AIF characteristic, thus focusing only on goal-oriented behaviour. Furthermore, and very relevant, the reward implementation in CarRacing-v0 might be not the best way to provide dAIF with rewards for proper preference learning.

\bibliographystyle{splncs04}
\bibliography{ref}

\appendix

\section{Model Parameters}
\label{appendix:parameters}
% General parameters
\begin{table}[hbtp!]
\centering
\label{tab:general}
\caption{General parameters}
\begin{tabular}{|l|l|l|}
\hline
\bf Parameter          & \bf Value  & \bf Description                                 \\ \hline
$N_{screens}$        & 8      & Size of the observation stack                           \\ \hline
$N_{colour}$          & 1      & Colour channels of the input image                       \\ \hline
$N_{height}$         & 42     & Height in pixels of the input image       \\ \hline
$N_{width}$          & 42     & Width in pixels of the input image        \\ \hline
$N_{actions}$        & 11     & Number of actions the agent can select from             \\ \hline
$N_{episodes}$       & 1000   & Number of episodes the model is trained for             \\ \hline
$N_{length\_episode}$ & 1000   & The maximum amount of time steps in an episode     \\ \hline
Freeze period & 50            & The amount of time steps the target network is frozen \\ & & before copying the parameters of the policy/value network                       \\ 
\hline
Batch size         & 250    & Number of items in a mini-batch                       \\
\hline
\end{tabular}
\end{table}

% DQN model
\begin{table}[hbtp!]
\centering
\label{tab:DQN}
\caption{DQN parameters}
\begin{tabular}{|l|l|l|}
\hline
\bf Parameter & \bf Value  & \bf Description  \\ \hline
Policy network & & Convolutional neural network which estimates Q-values given a state\\ 
 & & see Appendix B.\\ \hline
Target network  & & Copy of the Policy network which is updated after each freeze period \\
 & & see Appendix B.\\ \hline
$N_{hidden}$ & 512    & Number of hidden units in the policy and target network \\ \hline
$\lambda$    & 1e-5   & Learning rate    \\ \hline
$\gamma$     & 0.99   & Discount factor \\ \hline
$\epsilon$   & 0.15 $\rightarrow$ 0.05 & Probability of selecting a random action. (Starts as 0.15, \\
           & &  decreases linearly per episode with 0.00015 until a minimum of 0.05) \\ \hline
Memory capacity & 300000 & Number of transitions the replay memory can store   \\  \hline
\end{tabular}
\end{table}

% dAIF model
\begin{table}[hbtp!]
\centering
\label{tab:dAIF}
\caption{dAIF parameters}
\begin{tabular}{|l|l|l|}
\hline
\bf Parameter & \bf Value  & \bf Description  \\ \hline
Observation network & & VAE; see Appendix C.\\ \hline
Transition network & & Feed-forward neural network of shape:\\
&& (2$N_{latent}$ + 1) \text{ x } $N_{hidden}$ \text{ x } $N_{actions}$ \\ \hline
Policy network & & Feed-forward neural network of shape:\\
&& 2$N_{latent}$ \text{ x } $N_{hidden}$ \text{ x } $N_{actions}$\text{; with a}\\ 
&& \text{softmax function on the output}\\ \hline
Value network & & Feed-forward neural network of shape:\\
&& 2$N_{latent}$ \text{ x } $N_{hidden}$ \text{ x } $N_{actions}$ \\ \hline
Target network & & Copy of the Value network which is updated\\
&& after each freeze period \\ \hline
$N_{hidden}$  & 512    & Number of hidden units in the transition, \\ 
& & policy, and value network. \\ \hline
$N_{latent}$ & 128 & Size of the latent state \\ \hline
$\lambda_{transition}$     & 1e-3   & Learning rate of the transition network   \\ \hline
$\lambda_{policy}$ & 1e-4   & Learning rate of the policy network \\ \hline
$\lambda_{value}$ & 1e-5   & Learning rate of the value network \\ \hline
$\lambda_{VAE}$ & 5e-6   & Learning rate of the VAE \\ \hline
$\gamma$   & 12   & Precision parameter  \\ \hline
$\beta$    & 0.99 & Discount factor  \\ \hline
$\alpha$   & 18000 & $\frac{1}{\alpha}$ is multiplied with the VAE loss to scale its\\
& & size to that of the other term in the VFE \\ \hline
Memory capacity & 100000 & Number of transitions the replay memory can store \\  \hline
\end{tabular}
\end{table} 

\newpage
\section{DQN: Policy network}
\begin{table}[hbtp!]
    \centering
    \caption{Layers DQN policy network}
    \label{tab:DQN_net}
    \begin{tabular}{|c|c|c|c|c|c|}
        \hline
        Type & out channels & kernel & stride & input & output \\
        \hline
        conv & 64 & 4 & 2 &   (1, 8, 42, 42) & (1, 64, 20, 20)\\
        batchnorm &  &  &  &    & \\
        maxpool & - & 2 & 2 &   (1, 64, 20, 20) & (1, 64, 10, 10)\\
        relu &  &  &  &    & \\
        conv & 128 & 4 & 2 &   (1, 64, 10, 10) & (1, 128, 4, 4)\\
        batchnorm &  &  &  &    & \\
        maxpool & - & 2 & 2 &   (1, 128, 4, 4) & (1, 128, 2, 2)\\
        relu &  &  &  &    & \\
        conv & 256 & 2 & 2 &   (1, 128, 2, 2) & (1, 256, 1, 1)\\
        relu &  &  &  &    & \\
        dense & - & - & - &   256 & 512\\
        dense & - & - & - &   512 & 11\\
        \hline
    \end{tabular}

\end{table}

\newpage
\section{VAE}
%% VAE TABLE
\begin{table}[hbtp!]
    \centering
    \caption{VAE layers}
    \label{tab:vae}
    \begin{tabular}{|c|c|c|c|c|c|l}
        \cline{1-6}
        Type & out channels & kernel & stride & input & output \\
        \cline{1-6}
        conv & 32 & 4 & 2 &   (1, 8, 42, 42) & (1, 32, 20, 20) & \rdelim\}{14}{3mm}[Encoder] \\
        batchnorm &  &  &  &    & \\
        relu &  &  &  &    & \\
        conv & 32 & 4 & 2 &   (1, 32, 20, 20) & (1, 64, 9, 9)\\
        batchnorm &  &  &  &    & \\
        relu &  &  &  &    & \\
        conv & 128 & 5 & 2 &   (1, 64, 9, 9) & (1, 128, 3, 3)\\
        batchnorm &  &  &  &    & \\
        relu &  &  &  &    & \\
        conv & 256 & 3 & 2 &   (1, 128, 3, 3) & (1, 256, 1, 1)\\
        relu &  &  &  &    & \\
        
        dense & - & - & - &   256 & 128\\
        dense $\mu$ & - & - & - &   128 & 128\\
        dense log$\Sigma$ & - & - & - &   128 & 128\\
        dense & - & - & - &   128 & 128 & \rdelim\}{15}{3mm}[Decoder]\\
        dense & - & - & - &   128 & 256\\
        
        deconv & 128 & 3 & 2 &   (1, 256, 1, 1) & (1, 128, 3, 3) \\
        batchnorm &  &  &  &    & \\
        relu &  &  &  &    & \\
        deconv & 64 & 5 & 2 &   (1, 128, 3, 3) & (1, 64, 9, 9)\\
        batchnorm &  &  &  &    & \\
        relu &  &  &  &    & \\
        deconv & 32 & 4 & 2 &   (1, 64, 9, 9) & (1, 32, 20, 20)\\
        batchnorm &  &  &  &    & \\
        relu &  &  &  &    & \\
        deconv & 8 & 4 & 2 &   (1, 32, 20, 20) & (1, 8, 42, 42)\\
        batchnorm &  &  &  &    & \\
        relu & & & & &\\ % THIS IS "WRONG" (in terms of how one should make a VAE), BUT GAVE US THE BEST RESULTS
        sigmoid &  &  &  &  & \\
        \cline{1-6}
    \end{tabular}
\end{table}

\newpage

\section{Derivations}
Derivation for the EFE for a single time step:
\label{appendix:EFEderivation}
\begin{equation}
\begin{split}
G(s_k, o_k) &= \KL[q(s_k) \mid \mid p(s_k, o_k)] \\ 
&= \int q(s_k) \ln\frac{q(s_k)}{p(s_k,o_k)} \\
&=  \int q(s_k) \ln{q(s_k)} - \ln{p(s_k,o_k)}  \\
&= \int q(s_k) \ln{q(s_k)} - \ln{p(s_k|o_k)} - \ln{p(o_k)} \\
&\approx \int q(s_k) \ln{q(s_k)} - \ln{q(s_k|o_k)} - \ln{p(o_k)} \\
&\approx - \ln{p(o_k)} + \int q(s_k) \ln{q(s_k)} - \ln{q(s_k|o_k)} \\
&\approx - \ln{p(o_k)} + \int q(s_k) \ln{\frac{q(s_k)}{q(s_k|o_k)}} \\
&\approx - \ln{p(o_k)} + \KL[q(s_k) \mid \mid q(s_k|o_k)] \ \\
&\approx -r(o_k) + \KL[q(s_k) \mid \mid q(s_k|o_k)] \nonumber
\end{split}
\end{equation}

\newpage
\section{Average Reward over 100 episodes}
\label{appendix:avgreward}
\begin{figure} [hbtp!]
        \centering
        \includegraphics[width=1\linewidth]{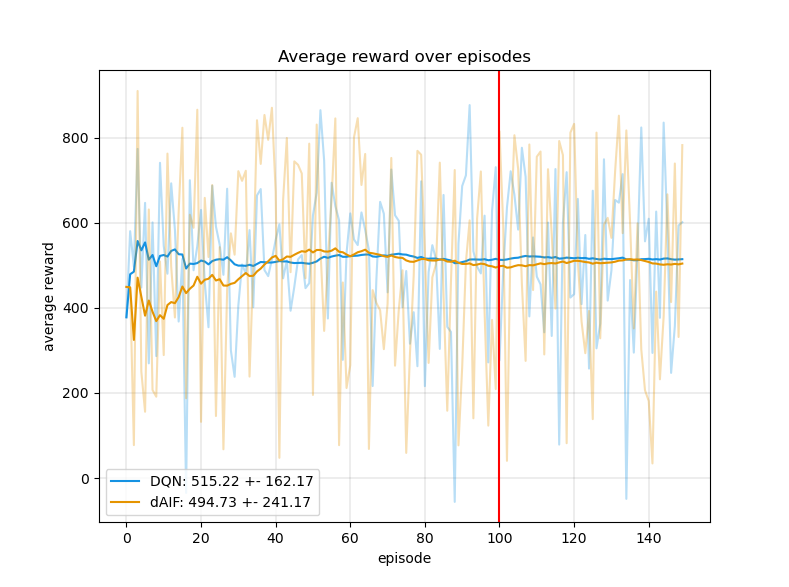}
        \caption{Average reward test over 100 episodes for DQN and dAIF. The bright lines show the mean over episodes. The transparent lines show the reward that was obtained in a particular episode.}
        \label{fig:avg_reward}
\end{figure}

\end{document}